\documentclass[oribibl]{llncs}

\usepackage{soul}
\usepackage{url}
\usepackage[small]{caption}
\usepackage{graphicx}
\usepackage{booktabs}
\newtheorem{examplealt}{Example}
\usepackage{hyperref}
\usepackage{caption}
\usepackage{graphicx}
\usepackage{amssymb}
\usepackage{amsmath}
\usepackage{latexsym}
\usepackage{color}
\usepackage{comment}
\usepackage{multicol}
\usepackage{amsmath}
\usepackage{amssymb}
\usepackage{graphicx}
\usepackage{soul}
\usepackage{epsfig}
\usepackage{color}
\usepackage{hhline}
\usepackage{enumerate}
\usepackage{times}
\usepackage{multirow}

\newcommand{\ignore}[1]{}
\newcommand{\boxtheorem}{\hfill $\Box$\\}
\newcommand{\nit}[1]{{\it #1}}

\newcommand{\mc}[1]{\mathcal{ #1}}

\newcommand{\msf}[1]{\mathsf{ #1}}

\newcommand{\e}{\mathbf{e}}
\newcommand{\C}[1]{\mathcal{C}}
\newcommand{\T}[1]{\mathcal{T}}

\newcommand{\shap}{{\sf Shap}}
\newcommand{\resp}{{\sf Resp}}

\usepackage[linesnumbered,algoruled,commentsnumbered]{algorithm2e}
\newcommand{\hrulealg}[0]{\vspace{1mm} \hrule \vspace{1mm}}

\begin{document}
\thispagestyle{empty}
\pagestyle{plain}

\title{Efficient Computation of Shap Explanation Scores for Neural Network Classifiers via Knowledge Compilation \ignore{\\ \underline{{\large REVISED, WITH APPENDIX}}}\vspace{-4mm}}

\author{{\bf Leopoldo Bertossi}\inst{1}\thanks{Member of the Millennium Institute for Foundational Research on Data (IMFD, Chile) \ leopoldo.bertossi@skema.edu} \and
{\bf Jorge E. Le\'on}\inst{2}\thanks{jorgleon@alumnos.uai.cl}}
\institute{SKEMA Business School, Montreal, Canada\\  \and Universidad Adolfo Ib\'a\~nez (UAI), Santiago, Chile}

\maketitle

\vspace{-6mm}
\begin{abstract}
The use of  \shap \ scores has become widespread in Explainable AI. However, their computation is in general intractable, in particular when done with a black-box classifier, such as neural network. Recent research has unveiled classes of open-box Boolean Circuit classifiers for which \shap \ can be computed efficiently. We show how to transform binary neural networks into those circuits for efficient \shap \ computation. We use logic-based knowledge compilation techniques. The performance gain is huge, as we show in the light of our experiments.  \vspace{-4mm}
\end{abstract}

\section{Introduction}
In recent years,  there has been a growing demand for methods to explain and interpret the results  from machine learning (ML) models.   Explanations come in different forms, and can be obtained through different approaches. A common one assigns {\em attribution scores} to the  features values associated to an input that goes through an ML-based model, to {\em quantify} their relevance for the obtained outcome. We concentrate on {\em local} scores, i.e. associated to a particular input, as opposed to a global score that indicated the overall relevance of a feature. We also concentrate on explanations for binary classification models that assign labels $0$ or $1$ to inputs.

A popular local score is \shap \ \cite{lundberg}, which is based on the Shapley value that was introduced in coalition game theory and practice \cite{S53,R88}. \ignore{Another attribution score that has been recently investigated in \cite{deem,RW21} is \resp,  the responsibility score \cite{halpernChockler} associated to  {\em actual causality} \cite{HP05}. In this work we concentrate on the \shap \ score, but the issues investigated here would be also interesting for  other scores.}
\shap \ scores can be computed with a black-box or an open-box model \cite{rudin}. With the former, we do not know or use its internal components, but only its input/output relation. This is the most common approach. In the latter case, we can have access to its internal structure and components, and we can use them for score computation.   It is common to consider neural-network-based models as black-box models, because  their internal gates and structure may  be difficult to understand or process when it comes to explaining classification outputs. However, a decision-tree model, due to its much simpler structure and use,  is considered to be open-box for the same purpose.

Even for binary classification models, the complexity of \shap \ computation is provably hard, actually $\#P$-hard for several kinds of  binary classification models, independently from whether the internal components of the model are used when computing \shap \ \cite{deem,aaai21JMLR,AAAI21}. However, there are classes of classifiers for which, using the model components and structure, the complexity of \shap \ computation can be brought down to polynomial time \cite{lundberg20,AAAI21,AAAI21Guy}.

 A polynomial time algorithm for \shap \ computation with  {\em deterministic and decomposable Boolean circuits} (dDBCs) was presented in \cite{AAAI21}. From this result, the tractability of \shap \ computation can be obtained for a
 variety of Boolean circuit-based classifiers and classifiers that can be represented as (or compiled into) them. In particular, this holds
for {\em Ordered Binary Decision Diagrams} (OBDDs) \cite{bryant},
decision trees,  and other established classification models that can be compiled  into (or treated as) OBDDs \cite{shi,darwicheEcai20,narodytska}. This applies, in particular, to {\em Sentential Decision Diagrams} (SDDs) \cite{darwicheSDDs} that form a convenient {\em knowledge compilation} target language \cite{darwicheKC,Broeck15}.
\ignore{In \cite{AAAI21Guy}, through a different approach, tractability of \shap \ computation was obtained for a collection of classifiers that intersect with that in \cite{AAAI21}.}

{\em In this work, we show how to use logic-based knowledge compilation techniques to attack, and -to the best of our knowledge- for the first time, the  important and timely problem of efficiently computing explanations scores in ML, which, without these techniques, would stay intractable.}

 More precisely, we concentrate on explicitly developing the compilation-based approach to the computation of \shap \ for {\em binary (or binarized) neural networks} (BNNs) \cite{hubara,qin,simons,narodytska}. For this, a BNN is transformed into a dDBC using techniques from {\em knowledge compilation} \cite{darwicheKC}, an area that investigates the transformation of (usually) propositional theories into an equivalent one with a canonical syntactic form that has some good computational properties, e.g. tractable model counting. The compilation may incur in a relatively high computational cost \cite{darwicheKC,darwicheJANCL}, but it may still be worth the effort when  a particular property is checked often, as is the case of explanations for the same BNN.

More specifically, we describe in detail how a BNN is first compiled into a propositional formula in Conjunctive Normal Form (CNF), which, in its turn, is compiled into an SDD, which is finally compiled into a dDBC.
Our method applies at some steps established transformations that are not commonly illustrated or discussed in the context of real applications, which we do here. The whole compilation path and  the application to \shap \ computation are new.
\ We show how \shap \ is computed on the resulting circuit via the efficient algorithm in \cite{AAAI21}.  This compilation is performed once, and  is independent from any input to the classifier. The final circuit can be used to compute \shap \ scores for different input entities.

We also make experimental comparisons of computation times between this open-box and circuit-based \shap \ computation, and that based directly on the BNN treated as a  black-box, i.e. using only its input/output relation. \
For our experiments, we consider real estate as an application domain, where house prices depend on certain features, which we appropriately binarize.\footnote{California Housing Prices dataset: https://www.kaggle.com/\\datasets/camnugent/california-housing-prices.} The problem consists in classifying property blocks, represented as entity records of thirteen feature values, as {\em high-value} or {\em low-value}, a binary classification problem for which a BNN is used.

To the best of our knowledge, our work is the first at using knowledge compilation techniques for efficiently computing \shap \ scores, and the first at  reporting experiments with the polynomial time algorithms for \shap \ computation on binary circuits.  \ We confirm that \shap \ computation via the dDBC vastly outperforms the direct \shap \ computation on the BNN. It is also the case that the scores obtained are fully  aligned, as expected since the dDBC represents the BNN. The same probability distribution associated to the Shapley value is used with all the models.

Compilation of  BNNs into OBDDs was done in \cite{shi,darwicheEcai20} for other purposes, not for \shap \ computation or any other kind of attribution score. \
In this work we concentrate only on explanations based on \shap \ scores. There are several other explanations mechanisms for ML-based classification and decision systems in general,  and also specific for neural networks. See \cite{fosca} and \cite{ras} for surveys.

This paper is structured as follows. Section \ref{sec:arte} contains background on \shap \ and Boolean circuits (BCs).  Section \ref{sec:trans} shows in detail, by means of a running example, the kind of  compilation of BNNs into dDBCs we use for the experiments. Section \ref{sec:comp} presents the experimental setup, and the results of our experiments with \shap \ computation.  In Section \ref{sec:con} we draw some conclusions. \vspace{-2mm}

\section{Preliminaries}\label{sec:arte}

\vspace{-2mm}In coalition game theory and its applications, the Shapley value is a established measure of the contribution of a player to a shared wealth that is modeled  as a game function. \
 Given a set of players $S$, and a game function $G: \mc{P}(S) \rightarrow \mathbb{R}$, mapping subsets of players to real numbers, the  Shapley value for a player $p \in S$ quantifies its contribution to $G$.
\ignore{% \begin{eqnarray}
% %\scriptsize
% \nit{Shapley}(S,G,p):= \hspace*{4cm}\label{eq:shapley}&& \\
% \sum_{s\subseteq
%   S \setminus \{p\}} \frac{|s|! (|S|-|s|-1)!}{|S|!}
% (G(s\cup \{p\})-G(s)).\hspace*{-0.8cm}&&\nonumber
% \end{eqnarray}
\begin{equation}
\nit{Shapley}(S,G,p):=
\sum_{s\subseteq
  S \setminus \{p\}} \frac{|s|! (|S|-|s|-1)!}{|S|!}
(G(s\cup \{p\})-G(s)).
\label{eq:shapley}
\end{equation}
It quantifies the contribution of  $p$  to $G$. } It emerges as the only measure that enjoys certain desired properties \cite{R88}.  In order to apply the Shapley value, one has to define an appropriate game function.

Now, consider a fixed entity $\e = \langle F_1(\e), \ldots, F_N(\e)\rangle$ subject to  classification. It has values  $F_i(\e)$ for  features in $\mc{F} = \{F_1, \ldots, F_N\}$. These values are $0$ or $1$ for binary features.  In \cite{lundberg,lundberg20}, the Shapley value  is applied with  $\mc{F}$ as the set of  players, and with the game function
$\mc{G}_\mathbf{e}(s) := \mathbb{E}(L(\mathbf{e'})~|~\mathbf{e'}_{\!s} = \mathbf{e}_s)$, giving rise to the \shap \ score. Here, $s \subseteq \mc{F}$, and $\mathbf{e}_s$ is the projection (or restriction) of $\e$ on (to) the  $s$. The label  function $L$ of the classifier assigns values $0$ or $1$. The $\e^\prime$ inside the expected value is an entity whose values coincide with those of $\e$ for the features in $s$.  \
For feature $F \in \mc{F}$:
\begin{eqnarray}
\mbox{\shap}(\mc{F},\mc{G}_\mathbf{e},F) =\sum_{s\subseteq
  \mc{F} \setminus \{F\}} \frac{|s|! (|\mc{F}|-|s|-1)!}{|\mc{F}|!}
~[\hspace*{40mm}&&\label{eq:SHAPfunction}\\
\mathbb{E}(L(\mathbf{e}')~|~\mathbf{e}'_{s\cup \{F\}} = \mathbf{e}_{s\cup \{F\}})- \mathbb{E}(L(\mathbf{e}')~|~\mathbf{e}'_s = \mathbf{e}_s)~].\hspace*{-5mm}&& \nonumber
\end{eqnarray}
The expected value assumes an underlying probability distribution on the entity population. \shap \ quantifies the contribution of feature value $F(\e)$ to the outcome label.

In order to compute \shap, we only need  function $L$, and none of the internal components of the classifier. Given that all possible subsets of features appear in its definition, \shap \ is bound to be hard to compute. Actually, for some classifiers, its computation may become $\#P$-hard \cite{AAAI21}. However, in \cite{AAAI21}, it is shown that \shap \ can be computed in polynomial time for every {\em deterministic and decomposable Boolean circuit} (dDBC) used as a classifier. The circuit's internal structure is used in the computation.

Figure \ref{fig:EjemplodDBC} shows  a Boolean circuit that can be used as a binary classifier, with binary features $x_1,x_2,x_3$, whose values are input at the bottom nodes, and then propagated upwards through the Boolean gates. The binary label is read off from the top node.
\ This circuit is {\em deterministic} in that, for every $\vee$-gate, at most one of its inputs is $1$ when the output is $1$. It is {\em decomposable} in that, for every $\wedge$-gate, the inputs do not share features.\ The dDBC in the Figure is also  {\em smooth}, in that  sub-circuits that feed a same $\vee$-gate share the same features. It has a {\em fan-in}   at most two, in that every $\wedge$-gate and $\vee$-gate have at most two inputs. We denote this subclass of dDBCs with dDBCSFi(2).

\begin{multicols}{2}
%\hfill
%\begin{minipage}[t]{0.4\textwidth}
%\phantom{oo}
\begin{center}
\includegraphics[width=4.2cm]{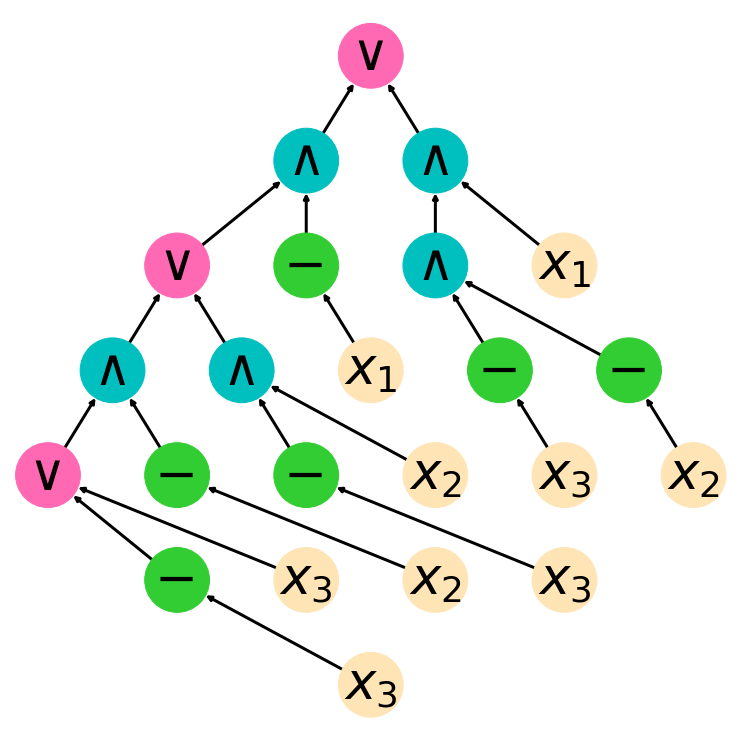}
\vspace{-2mm}
\captionof{figure}{A dDBC.}\label{fig:EjemplodDBC}

\phantom{oo}

\includegraphics[width=4.5cm]{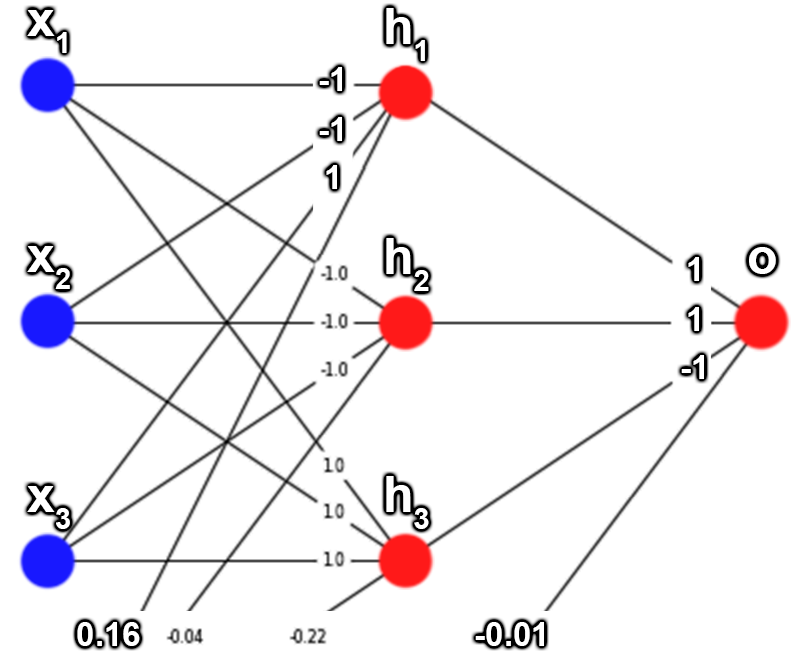}
\vspace{-2mm}
\captionof{figure}{A BNN.}\label{fig:EjBNN}
\end{center}
%\end{minipage}
\end{multicols}

\vspace{-2mm}
More specifically, in \cite{AAAI21} it is established that \shap \ can be computed in polynomial time for dDBCSFi(2)-classifiers, assuming that the underlying probability distribution is the uniform,  $P^{\msf{u}}$, or the product distribution, $P^\times$. They are as follows for binary  features: \
$P^{\msf{u}}(\e) := \frac{1}{2^N}$ and
$P^\times(\e) := \Pi_{i=1}^N p_i(F_i(\e))$,
where  $p_i(v)$ is the probability of value $v \in \{0,1\}$ for feature $F_i$.

%%%%%%%%%%%%%%%%%%%%%%%%

%%%%%%%%%%%%%%%%%%%%%%%%

\vspace{-2mm}
\section{Compiling BNNs into dDBCs}\label{sec:trans}

\vspace{-1mm}In order to compute \shap \ with a BNN, we convert the latter into a dDBC, on which \shap \ scores will be computed  with  the polynomial time algorithm  in \cite{AAAI21}. \ The transformation goes along the the following path that we describe in this section:
\begin{equation}
\stackrel{\mbox{BNN}}{\mbox{\phantom{oo}}} \ \ \  \stackrel{\longmapsto}{\mbox{{\footnotesize (a)}}} \ \ \ \stackrel{\mbox{CNF}}{\mbox{\phantom{oo}}} \ \ \  \stackrel{\longmapsto}{\mbox{{\footnotesize (b)}}} \ \ \  \stackrel{\mbox{SDD}}{\mbox{\phantom{oo}}} \ \ \  \stackrel{\longmapsto}{\mbox{\footnotesize (c)}} \ \ \  \stackrel{\mbox{dDBC}}{\mbox{\phantom{oo}}} \label{eq:path}
\end{equation}

 \vspace{-2mm}A BNN can be converted into a CNF formula  \cite{shihcnf,narodytska}, which, in its turn, can be converted into an SDD \cite{darwicheSDDs,decSDD}. It is also known that SDDs can be compiled into a formula in d-DNNF (deterministic and decomposable negated normal form) \cite{darwicheKC}, which forms a subclass of dDBCs. More precisely, the resulting dDBC in (\ref{eq:path}) is finally compiled in polynomial time into a dDBCSFi(2).

Some of the steps in (\ref{eq:path}) may not be polynomial-time transformations, which we will discuss in more technical terms later in this section. \  However, we can claim at this stage that: (a) Any exponential cost of a transformation is kept under control by a usually small parameter. \ (b) The resulting  dDBCSFi(2) is meant to be used multiple times, to explain different and multiple outcomes; and then, it may be worth taking a one-time, relatively high transformation cost.  \
A good reason for  our transformation path is the availability of implementations we can take advantage of.\footnote{The path in (\ref{eq:path}) is not the only way to obtain a dDBC. For example, \cite{shi} describe  a conversion of BNNs into OBDDs, which can also be used to obtain  dDBCs. However, the asymptotic time complexity is basically the same.}

We will describe, explain and illustrate the conversion path (\ref{eq:path}) by means of a running example with a simple BNN, which is not the BNN used for our experiments. For them,  we used a  BNN with one hidden layer with 13 gates.

\vspace{-1mm}
\begin{examplealt} \label{ex;ex1} \em The BNN in Figure \ref{fig:EjBNN} has hidden neuron gates $h_1, h_2,h_3$, an output gate $o$, and three input gates, $x_1, x_2,x_3$, that receive  binary values.  The latter represent, together,  an input entity $\bar{x} =\langle x_1,x_2,x_3\rangle$ that is being classified by means of a label returned by $o$. \
Each  gate $g$ is activated by means of a {\em step function} \ $\phi_g(\bar{i})$ \ of the form:
 \vspace{-2mm}
\begin{equation}
\nit{sp}(\bar{w}_g\bullet \bar{i}+b_g) \ := \left\{\begin{array}{rl}
\hspace{-3mm}1  & \hspace*{0mm} \mbox{ if } \ \bar{w}_g\bullet \bar{i}+b_g \geq 0, \label{eq:thr}\\
-1 & \ \mbox{otherwise and $g$ is hidden},\\
0 & \mbox{ otherwise and $g$ is output},   \end{array} \right.
\end{equation}

\vspace{-2mm}
\noindent which is parameterized by a vector of binary weights $\bar{w}_g$ and a real-valued constant bias $b_g$.\footnote{We could also used binarized {\em sigmoid} and {\em softmax} functions.} The $\bullet$  is the inner vector product. For technical, non-essential reasons, for a hidden gate, $g$, we  use $1$ and $-1$, instead of $1$ and $0$, in $\bar{w}_g$   and outputs. Similarly, $\bar{x} \in \{-1,1\}^3$. Furthermore, we assume we have a single output gate, for which the activation function does return $1$ or $0$,  for {\em true} or {\em false}, respectively.

For example,  $h_1$ is  {\em true}, i.e. outputs $1$, for an input $\bar{x} = (x_1,x_2,x_3)$ iff $\bar{w}_{h_1} \! \bullet \bar{x} + b_{h_1} = (-1) \times x_1 + (-1)  \times x_2 + 1 \times x_3 + 0.16 \geq 0$.  Otherwise, $h_1$ is {\em false}, i.e. it returns $-1$. \
Similarly, output gate $o$ is {\em true}, i.e. returns label $1$ for a binary input $\bar{h} = (h_1,h_3,h_3)$ iff $\bar{w}_o \bullet \bar{h} = 1 \times h_1 + 1 \times h_2 + (-1) \times h_3 - 0.01 \geq 0$, and $0$ otherwise.  \boxtheorem
\end{examplealt}

\vspace{-6mm}
The first step, (a) in (\ref{eq:path}),  represents the BNN as a CNF formula, i.e. as a conjunction of disjunctions of {\em literals}, i.e. atomic formulas or their negations.
\ignore{ For this, we adapt the approach  in \cite{narodytska}, in their case, to verify properties of BNNs. Contrary to them,  we avoid  the use of auxiliary variables since their posterior elimination conflicts with our need for determinism.}

Each gate of the BNN is  represented by a propositional formula, initially not necessarily in CNF, which, in its turn,   is used as one of the inputs to gates next to the right.   In this way, we eventually obtain a defining formula for the output gate. The formula is converted into CNF.   The participating propositional variables are logically treated as {\em true} or {\em false}, even if they take numerical values $1$ or $-1$, resp.

\vspace{-2mm}
\subsection{Representing BNNs as Formulas in CNF}
\label{BNNtoCNF}

\vspace{-1mm}
Our conversion of the BNN into a CNF formula is inspired by a technique introduced in \cite{narodytska}, in their case, to verify properties of BNNs. In our case, the NN is fully binarized in that inputs, parameters (other than bias), and outputs are always binary, whereas they may have real values as parameters and outputs. Accordingly, they have to binarize values along the transformation process. They also start producing logical constraints that are later transformed into CNF formulas. Furthermore, \cite{narodytska} introduces auxiliary variables during and at the end of the transformation. With them, in our case, such a BC could not be used for \shap \ computation. Furthermore, the elimination of auxiliary variables, say via {\em variable forgetting} \cite{forSDD},  could harm the determinism of the final circuit.
\ In the following we describe a transformation that avoids introducing auxiliary variables.\footnote{At this point is  where using $1,-1$ in the BNN instead of $1,0$ becomes useful.} However, before describing the method in general, we give an example, to convey the main ideas and intuitions.

\begin{examplealt} \label{ex;ex2} \em (example \ref{ex;ex1} cont.) Consider gate $h_1$, with parameters  $\bar{w} = \langle -1,-1,1\rangle$ and $b=0.16$, and input $\bar{i} = \langle x_1,x_2,x_3 \rangle$. An input $x_j$ is said to be {\em conveniently instantiated} if it has the same sign as $w_j$, and then, contributing to having a larger number on the LHS of the comparison in (\ref{eq:thr}). E.g., this is the case of \ $x_1 = -1$. \ In order to represent as a  propositional formula its output variable, also denoted with $h_1$, we  first compute the number, $d$, of conveniently instantiated inputs that are necessary and sufficient to make the LHS of the comparison in  (\ref{eq:thr}) greater than or equal to $0$. This is the (only) case when $h_1$ becomes  {\em true}; otherwise, it is {\em false}. This number can be computed in general by \cite{narodytska}:
\begin{equation}
    d = \left \lceil (-b + \sum_{j=1}^{|\bar{i}|} w_{j})/2 \right \rceil + \mbox{\em \# of negative weights in } \bar{w}. \label{eq:d}
    \end{equation}

For $h_1$, with 2 negative weights: \ $d(h_1)= \left \lceil (-0.16 + (-1 -1 + 1))/2 \right \rceil + 2 = 2$. \
With this, we can impose conditions on two input variables with the right sign at a time, considering all possible convenient pairs. For $h_1$ we obtain its condition to be true: \vspace{-2mm} \begin{equation}h_1 \ \longleftrightarrow \ (-x_1 \wedge -x_2) \vee (-x_1 \wedge x_3) \vee (-x_2 \wedge x_3). \label{eq:dnf}\end{equation}

\vspace{-3mm}
This DNF formula is  directly obtained -and just to convey the intuition- from considering all possible convenient pairs (which is already better that trying all cases of three variables at a time). However,  the general iterative method presented \ignore{earlier}later in this subsection, is more expedite and compact than simply listing all possible cases; and  still uses the  number of convenient inputs. Using this general algorithm, we obtain, instead of (\ref{eq:dnf}),  this equivalent  formula defining $h_1$:\vspace{-2mm}
\begin{equation}
    h_1 \ \longleftrightarrow \ (x_3 \wedge (-x_2 \vee -x_1)) \vee (-x_2 \wedge -x_1).
    \label{eq:h1encoded}
\end{equation}

\vspace{-2mm}
Similarly, we obtain defining formulas for $h_2$, $h_3$, and $o$: \ (for all of them, $d=2$)\vspace{-2mm}
\begin{eqnarray}
h_2 &\longleftrightarrow& (-x_3 \wedge (-x_2 \vee -x_1)) \vee (-x_2 \wedge -x_1), \nonumber\\
        h_3 &\longleftrightarrow& (x_3 \wedge (x_2 \vee x_1)) \vee (x_2 \wedge x_1), \nonumber\\
           o &\longleftrightarrow& (-h_3 \wedge (h_2 \vee h_1)) \vee (h_2 \wedge h_1).\label{eq:oencoded}
\end{eqnarray}

\vspace{-2mm}
Replacing the definitions of $h_1,h_2,h_3$ into (\ref{eq:oencoded}), we finally obtain:\vspace{-2mm}
\begin{eqnarray}
 o &\longleftrightarrow & (-[(x_3 \wedge (x_2 \vee x_1)) \vee (x_2 \wedge x_1)] \wedge ([(-x_3 \wedge (-x_2 \vee -x_1)) \vee (-x_2 \wedge -x_1)] \nonumber \\
    && \vee [(x_3 \wedge (-x_2 \vee -x_1)) \vee (-x_2 \wedge -x_1)])) \vee ([(-x_3 \wedge (-x_2 \vee -x_1)) \vee  \nonumber \\
    && (-x_2 \wedge -x_1)] \wedge \ [(x_3 \wedge (-x_2 \vee -x_1)) \vee (-x_2 \wedge -x_1)]).
    \label{eq:finaloencoded}
\end{eqnarray}

\vspace{-2mm}The final part of step (a) in path (\ref{eq:path}), requires transforming this formula into CNF. In this example, it can be taken straightforwardly into CNF. For our experiments,  we  implemented and used the general algorithm presented right after this example.  It guarantees that the generated CNF formula does not grow unnecessarily by eliminating some  redundancies along the process. \ignore{ (e.g. we tranform $(x_1 \vee x_2 \vee x_3) \wedge (x_1 \vee x_2) \wedge \ldots$ into $(x_1 \vee x_2) \wedge \ldots$), but these simplifications are not thorough, as they are only meant to not run out of memory.}
The resulting CNF formula is, in its turn, simplified into a shorter and simpler new CNF formula by means of the
 {\em Confer} SAT solver \cite{riss}. For this example, the simplified CNF formula is as follows:\vspace{-2mm}
\begin{equation}
o \ \longleftrightarrow \ (-x_{1} \vee -x_{2}) \wedge (-x_{1} \vee -x_{3}) \wedge (-x_{2} \vee -x_{3}). \label{eq:cnf}
\end{equation}

\vspace{-3mm}
Having a CNF formula will  be convenient for the next steps along path (\ref{eq:path}).
\boxtheorem \end{examplealt}

\vspace{-4mm}In more general terms, consider a BNN with $L$ layers, numbered with $Z \in [L] :=\{1,\ldots,L\}$. W.l.o.g., we may assume all layers have $M$ neurons (a.k.a. gates), except for the last layer that has a single, output neuron. We also assume that every neuron receives an input from every neuron at the preceding layer. Accordingly, each neuron at the first layer receives the same  binary input $\bar{i}_1 = \langle x_1,\ldots,x_N\rangle$ containing the values for the propositional input variables  for the BNN.  Every neuron at a layer $Z$ to the right receives the same binary input $\bar{i}_Z = \langle i_1, \ldots, i_M\rangle$ formed by the output values from the $M$ neurons at layer $Z-1$. \ Variables $x_1, \ldots, x_N$ are the only variables that will appear in the final CNF representing the BNN.\footnote{We say ``a CNF" meaning ``a formula in CNF". Similarly in plural.}

To convert the BNN into a representing CNF, we iteratively convert every neuron into a CNF, layerwise and from input to output (left to right). The CNFs representing neurons at a given layer $Z$ are used to build all the CNFs representing the  neurons at layer $Z+1$.

Now, for each neuron $g$, at a layer $Z$, the representing CNF, $\varphi^g$,  is constructed using a matrix-like structure $M^g$ with dimension $M \times d_g$, where $M$ is the number of inputs to $g$ (and $N$ for the first layer), and $d_g$ is computed as in (\ref{eq:d}), i.e. the number of inputs to conveniently instantiate to get output $1$.  Formula $\varphi^g$ represents $g$'s activation function $\nit{sp}(\bar{w}_g\bullet \bar{i}+b_g)$. \ The entries $c_{ij}$ of $M^g$
contain  terms of the form $w_k \cdot i_k$, which are not interpreted as numbers, but as propositions, namely $i_k$ if $w_k=1$, and $\neg i_k$ if $w_k = -1$ (we recall that $i_k$ is the $k$-th binary input to $g$, and $w_k$ is the associated weight).

Each $M^{g}$ is iteratively constructed in a row-wise manner starting from the top, and then column-wise from left to right, as follows: \ (in it, the $c_{ik}$ are entries already created in the same matrix)

\vspace*{-3mm}
\begin{equation}
    \scriptsize
    M^{g} =
    % \begingroup
    \setlength\arraycolsep{3.2pt}
    \begin{bmatrix}
    w_{1}\cdot i_{1} & \nit{false} & \nit{false} & \dots & \nit{false} \\[1.0ex]
    \begin{matrix}w_{2}\cdot i_{2}\\[-0.50ex]\vee c_{11}\end{matrix} & \begin{matrix}w_{2}\cdot i_{2}\\[-0.50ex]\wedge c_{11}\end{matrix} & \nit{false} & \dots & \nit{false} \\[2.5ex]
    \begin{matrix}w_{3}\cdot i_{3}\\[-0.50ex]\vee c_{21}\end{matrix} & \begin{matrix}(w_{3}\cdot i_{3}\\[-0.50ex]\wedge c_{21})\\[-0.50ex]\vee c_{22}\end{matrix} & \begin{matrix}w_{3}\cdot i_{3}\\[-0.50ex]\wedge c_{22}\end{matrix} & \dots & \nit{false} \\[0.5ex]
    \dots & \dots & \dots & \dots & \dots \\[0.5ex]
    \begin{matrix}w_M\cdot i_M \vee \\[-0.50ex] c_{(M-1)1} \end{matrix} & \begin{matrix}(w_M\cdot i_M\\[-0.50ex]\wedge c_{(M-1)1})\\[-0.50ex]\vee c_{(M-1)2}\end{matrix} & \begin{matrix}(w_M\cdot i_M\\[-0.50ex]\wedge c_{(M-1)2})\\[-0.50ex]\vee c_{(M-1)3}\end{matrix} & \dots & \begin{matrix}\underline{(w_M\cdot i_M \wedge}\\[-0.50ex]\underline{c_{(M-1)(d_g-1)})}\\[-0.50ex]\underline{\vee c_{(M-1)d_g}}\end{matrix} \\
    \end{bmatrix}
    \normalsize
    \label{eq:matrixM}
\end{equation}

The $k$-th row represents the first $k \in [M]$ inputs considered for the encodings, and each column, the threshold $t \in [d_g]$ to surpass, meaning that at least $t$ inputs should be instantiated conveniently. For every component $c_{k,t}$with $k < t$, the threshold cannot be reached, which makes every component in the upper-right triangle  $\nit{false}$.

The propositional formula of interest, namely the one that represents neuron $g$ and  will be passed over as an ``input" to the next layer to the right, is the bottom-right most, $c_{Md_g}$ (underlined).
Notice that it is not necessarily a CNF; nor does the construction of $M^g$ requires using CNFs. It is also clear that, as we construct the components of matrix $M^g$, they become  formulas that keep growing in size.   Accordingly, before passing over this formula, it is converted into a CNF $\varphi^g$ that has also been simplified by means of a SAT solver (this, at least experimentally, considerably reduces the size of the CNF). The vector $\langle \varphi^{g_1}, \ldots, \varphi^{g_M}\rangle$ becomes the input for the construction of the matrices $M^{g\prime}$, for neurons $g\prime$ in layer $Z+1$. \ Reducing the sizes of these formulas is important because the construction of  $M^{g\prime}$ will make the the formula sizes grow furher.   %\red{See Figure \ref{fig:BNNaCNF}. }

\begin{examplealt} \em (ex. \ref{ex;ex2} cont.) \ Let us  encode neuron $h_1$ using the matrix-based construction. Since  $d_{h_1} = 2$, and it has $3$ inputs, matrix $M^{h_1}$ will have dimension $3\times2$.  Here, $\bar{w}_{h_1} = \langle -1,-1,1\rangle$ and $\bar{i}_{h_1} = \langle x_1,x_2,x_3 \rangle$. \ Accordingly,  $M_{h_1}$ has the following structure:

\begin{center}
    $\begin{bmatrix}
    w_{1}\cdot i_{1} & \nit{false} \\
    w_{2}\cdot i_{2} \vee c_{11} & w_{2}\cdot i_{2} \wedge c_{11} \\
    \phantom{-}w_{3}\cdot i_{3} \vee c_{21}\phantom{-} & \phantom{-}\underline{(w_{3}\cdot i_{3} \wedge c_{21}) \vee c_{22}}\phantom{-} \\
    \end{bmatrix}$
\end{center}
Replacing in its components the corresponding values, we obtain:

\begin{center}
    $\begin{bmatrix}
    -x_{1} & \nit{false} \\
    -x_{2} \vee -x_{1} & -x_{2} \wedge -x_{1} \\
    \phantom{-}x_{3} \vee -x_{2} \vee -x_{1}\phantom{-} & \phantom{-}\underline{(x_{3} \wedge (-x_{2} \vee -x_{1})) \vee (-x_{2} \wedge -x_{1})}\phantom{-} \\
    \end{bmatrix}$
\end{center}
The highlighted formula coincides with that in (\ref{eq:h1encoded}). \boxtheorem
\end{examplealt}

\vspace{-4mm}
In our implementation, and this is a matter of  choice and convenience, it turns out that  each component of $M^g$ is transformed right away into a simplified CNF before being used to build the new components. This is not essential, in that we could, in principle, use (simplified) propositional formulas of any format all long the process,  but making sure that the final formula representing the whole BNN is in CNF.  \ Notice that all the $M^{g}$ matrices for a same layer $Z \in {L}$ can be generated in parallel and without interaction. Their encodings do not influence each other.  With this construction, no auxiliary propositional  variables other that those for the initial inputs are created.

Departing from  \cite{narodytska}, our use of the $M^{g}$ arrays  helps us directly build (and work with) CNF formulas without auxiliary variables all along the computation. The final CNF formula, which then contains only the input variables for the BNN,  is eventually transformed into a dDBC. \ The use of a SAT solver for simplification of formulas is less of a problem in \cite{narodytska} due to the use of auxiliary variables. \ Clearly, our simplification steps make us incur in an extra computational cost. However, it helps us mitigate the exponential growth of  the CNFs generated during the transformation of the BNN into the representing CNF.

Overall, and in the worst case that no formula simplifications are achieved, having still used the SAT solver, the time complexity of building the final CNF is exponential in the initial input. This is due to the growth of the formulas along the process. The number of operations in which they are involved in the matrices construction is quadratic.

\vspace{-3mm}\subsection{Building an SDD Along the Way}
\label{sec:SDD}

\vspace{-1mm}Following with step (b) along path (\ref{eq:path}), the resulting CNF formula is  transformed into a {\em Sentential Decision Diagram} (SDD) \cite{darwicheSDDs,Broeck15}, which, as a particular kind of {\em decision diagram} \cite{bollig}, is  a directed acyclic graph. So as the popular OBDDs \cite{bryant}, that SDDs generalize, they can be used to represent general Boolean formulas, in particular, propositional formulas (but without necessarily being {\em per se}  propositional formulas).
%\vspace{-2mm}

\begin{examplealt} \ \label{ex:sdd} \em (example \ref{ex;ex2} cont.) \
Figure \ref{fig:EjemploSDD}(a) shows an SDD, $\mc{S}$, representing our CNF formula on the RHS of (\ref{eq:cnf}). An SDD has different kinds of nodes.
Those represented with encircled numbers  are {\em decision nodes} \cite{Broeck15}, e.g.
\raisebox{.5pt}{\textcircled{\raisebox{-.9pt} {{\footnotesize $1$}}}} and \raisebox{.5pt}{\textcircled{\raisebox{-.9pt} {{\footnotesize $3$}}}}, that consider alternatives for the inputs (in essence, disjunctions). There are also nodes called {\em elements}. They are labeled with
constructs of the form $[\ell_1|\ell_2]$, where $\ell_1, \ell_2$, called the {\em prime} and the {\em sub}, resp., are
Boolean literals, e.g.  $x_1$ and  $\neg x_2$, including
$\top$ and $\bot$, for $1$ or $0$, resp. E.g.  $[\neg x_2| \top]$ is one of them. The {\em sub} can also be a pointer, $\bullet$, with an edge to a decision node. \ $[\ell_1|\ell_2]$ represents two conditions that have to be satisfied simultaneously  (in essence, a conjunction). An element without $\bullet$ is a  {\em terminal}.  (See \cite{bova,nakamura}
for a precise  definition of SDD.)

\vspace{-7mm}
\begin{figure}%[H]
\centering
\includegraphics[height=5.0cm]{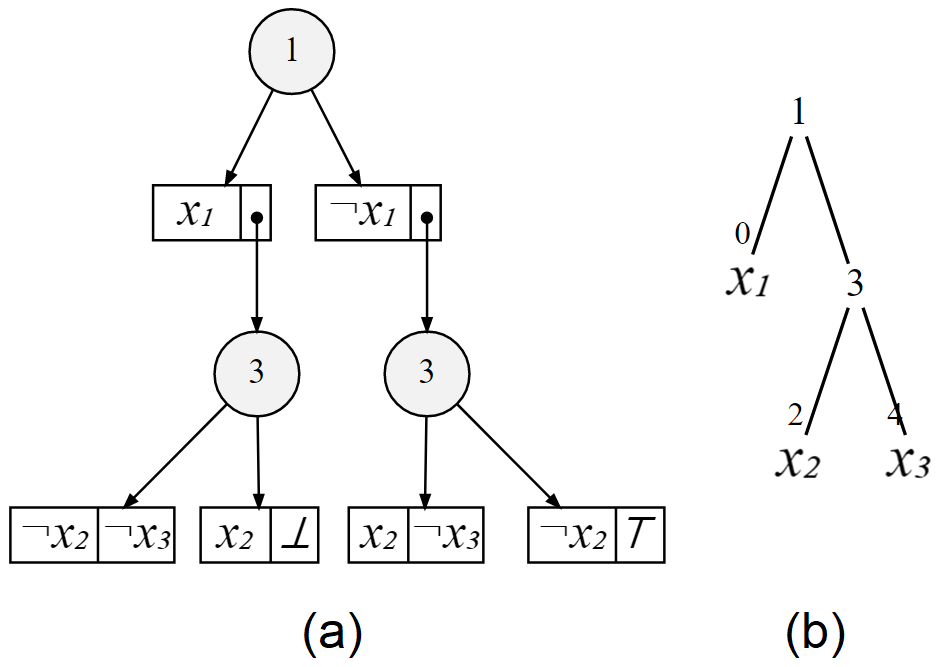}\vspace{-1mm}
\vspace{-2mm}\caption{ An SDD (a) and a vtree (b).}

\vspace{-2mm}\label{fig:EjemploSDD}\label{fig:Ejvtree}\vspace{-3mm}
\end{figure}

%\vspace{-2mm}
\noindent  \ An SDD represents (or defines) a total Boolean function $F_{\mc{S}}\!\!: \langle x_1,x_2,x_3\rangle \in \{0,1\}^3 \mapsto \{0,1\}$. For example,  $F_{\mc{S}}(0,1,1)$ is evaluated by following the graph downwards. Since $x_1 =0$, we descent to the right; next via node \raisebox{.5pt}{\textcircled{\raisebox{-.9pt} {{\footnotesize $3$}}}} underneath, with $x_2 = 1$, we reach the instantiated leaf node labeled with $[1|0]$, a ``conjunction", with the second component due to $x_3 =1$.  We obtain  $F_{\mc{S}}(0,1,1) = 0$. \boxtheorem
\end{examplealt}

\vspace{-4mm}
In SDDs, the orders of occurrence of variables in the diagram must be compliant with a so-called {\em vtree} (for ``variable tree").\footnote{Extending OBDDs, whose vtrees make variables in a path always appear in the same order. This generalization makes SDDs much more succinct than OBDDs  \cite{Broeck15,bova,bollig}.} The connection between a vtree and an SDD refers to the compatibility between the partitions $[\nit{prime}|\nit{sub}]$ and the tree structure (see Example \ref{ex:vtree} below). \  Depending on the chosen vtree, substructures of an SDD can be better reused when representing a Boolean function, e.g. a propositional formula, which becomes  important  to obtain a compact representation.  SDDs can easily be combined via propositional operations, resulting in a new SDD \cite{darwicheSDDs}.

A vtree for a set of variables $\mc{V}$ is binary tree that is full, i.e. every node has $0$ or $2$ children, and
ordered, i.e. the children of a node are totally ordered, and there is a bijection between the set of  leaves and $\mc{V}$ \cite{bova}.

\vspace{-2mm}
\begin{examplealt} \label{ex:vtree} \em (example \ref{ex:sdd} cont.) \ Figure \ref{fig:Ejvtree}(b) shows a vtree, $\mc{T}$,  for $\mc{V} = \{ x_1, x_2, x_3\}$.  Its leaves, $0,2,4$, show their associated variables in $\mc{V}$. \ The SDD $\mc{S}$ in Figure \ref{fig:EjemploSDD}(a) is compatible with $\mc{T}$. \  Intuitively, the variables at $\mc{S}$'s terminals, when they go upwards through decision nodes \raisebox{.5pt}{\textcircled{\raisebox{-.9pt} {{\footnotesize $n$}}}}, also go upwards through the corresponding nodes $n$ in $\mc{T}$. (\ignore{C.f.}See \cite{bova,nakamura,bollig} for a precise, recursive definition.)

The SDD $S$ can be straightforwardly represented as a propositional formula by interpreting decision points as disjunctions, and elements as conjunctions, obtaining
\ $ [x_1 \wedge ((-x_2 \wedge -x_3) \vee (x_2 \wedge \bot))] \vee
[-x_1 \wedge ((x_2 \wedge -x_3) \vee (-x_2 \wedge \top))]$, which is logically equivalent to the formula on the RHS of (\ref{eq:cnf}) that represents our BNN. \ignore{Accordingly, the BNN is represented by the SDD  in Figure \ref{fig:EjemploSDD}(a).}
\boxtheorem
\end{examplealt}

\vspace{-5mm}
For the running example and experiments, we used the {\em PySDD} system \cite{PySDD}: Given a CNF formula $\psi$, it computes an associated vtree and a compliant SDD, both optimized in size \cite{choiapply,SDDmanual}. \
This compilation step, the theoretically most expensive along path (\ref{eq:path}), takes exponential space and time only in $\nit{TW}(\psi)$, the {\em tree-width}   of the {\em primal graph} $\mc{G}$ associated to $\psi$ \cite{darwicheSDDs,decSDD}. $\mc{G}$ contains the variables as nodes, and undirected edges between any of them when they appear in a same clause. \ The tree-width measures how close the graph is to being a tree. \  This is a positive {\em fixed-parameter tractability} result \cite{flum}, in that $\nit{TW}(\psi)$ is in general smaller than $|\psi|$.
\ For example, the graph $\mc{G}$ for the formula $\psi$ on the RHS of (\ref{eq:cnf}) has  $x_1, x_2, x_3$ as nodes, and edges between any pair of variables, which makes  $\mc{G}$ a complete graph. Since every complete graph has a tree-width equal to the number of nodes minus one, we have $\nit{TW}(\mc{\psi}) =2$. \ Overall, this step in the transformation process has a time complexity that, in the worst case, is exponential in the size of the tree-width of the input CNF.

\vspace{-2mm}
\subsection{The Final dDBC}
\label{sec:dDBC}
\vspace{-1mm}
Our final  dDBC is obtained from the resulting SDD:  An  SDD  corresponds to  a d-DNNF Boolean circuit, for which decomposability and determinism hold, and has only variables as inputs to negation gates \cite{darwicheSDDs}.  And d-DNNFs are also dDBCs. \ Accordingly, this step of the whole transformation is basically for free, or better, linear in the size of the SDD if we locally convert decision nodes into disjunctions, and elements into conjunctions (see Example \ref{ex:vtree}).

%\vspace{-5mm}

\begin{algorithm}[t]
\caption{ \ From dDBC to dDBCSFi(2)}\label{alg:SDD2dDBC}
\footnotesize
\SetKwInOut{Input}{Input}\SetKwInOut{Output}{Output}\SetKwProg{uFunction}{function}{}{}\SetKwProg{For}{for}{}{}
\Input{Original $\nit{dDBC}$ \ (starting from root node).}
\Output{A $\nit{dDBCSFi(2)}$ equivalent to the given $\nit{dDBC}$.}
\hrulealg
\uFunction{\upshape FIX\_NODE($\nit{dDBC\_node}$)}{
    \uIf{\upshape $\nit{dDBC\_node}$ is a disjunction}{
        $\nit{c_{new}}$ = $\nit{false}$\\
        \For{\upshape each subcircuit $sc$ in $\nit{dDBC\_node}$}{
            $\nit{sc_{fixed}}$ = FIX\_NODE($sc$)\\
            \uIf{\upshape $\nit{sc_{fixed}}$ is a $\nit{true}$ value or is equal to $\nit{\neg c_{new}}$}{
                \Return $\nit{true}$
            }
            \uElseIf{\upshape $\nit{sc_{fixed}}$ is not a $\nit{false}$ value}{
                \For{\upshape each variable $v$ in $\nit{c_{new}}$ and not in $\nit{sc_{fixed}}$}{
                    $\nit{sc_{fixed}}$ = $\nit{sc_{fixed}} \wedge (v \vee \neg v)$
                }
                \For{\upshape each variable $v$ in $\nit{sc_{fixed}}$ and not in $\nit{c_{new}}$}{
                    $\nit{c_{new}}$ = $\nit{c_{new}} \wedge (v \vee \neg v)$
                }
                $\nit{c_{new}}$ = $\nit{c_{new}} \vee \nit{sc_{fixed}}$
            }
        }
        \Return $\nit{c_{new}}$
    }
    \uElseIf{\upshape $\nit{dDBC\_node}$ is a conjunction}{
        $\nit{c_{new}}$ = $\nit{true}$\\
        \For{\upshape each subcircuit $sc$ in $\nit{dDBC\_node}$}{
            $\nit{sc_{fixed}}$ = FIX\_NODE($sc$)\\
            \uIf{\upshape $\nit{sc_{fixed}}$ is a $\nit{false}$ value or is equal to $\nit{\neg c_{new}}$}{
                \Return $\nit{false}$
            }
            \uElseIf{\upshape $\nit{sc_{fixed}}$ is not a $\nit{true}$ value}{
                $\nit{c_{new}}$ = $\nit{c_{new}} \wedge \nit{sc_{fixed}}$
            }
        }
        \Return $\nit{c_{new}}$
    }
    \uElseIf{\upshape $\nit{dDBC\_node}$ is a negation}{
        \Return $\neg$FIX\_NODE($\neg\nit{dDBC\_node}$)
    }
    \uElse{
        \Return $\nit{dDBC\_node}$
    }
}
$\nit{dDBCSFi(2)}$ = FIX\_NODE($\nit{root\_node}$)
\end{algorithm}

The algorithm in \cite{aaai21JMLR} for efficient \shap \ computation needs the dDBC to be a \linebreak dDBCSFi(2). To obtain the latter, we use the transformation    Algorithm \ref{alg:SDD2dDBC} below, which is based on \cite[sec. 3.1.2]{aaai21JMLR}. \ In a bottom-up fashion,  fan-in 2 is achieved by  rewriting every $\wedge$-gate (resp., and $\vee$-gate) of fan-in $m > 2$ with a chain of $m-1$ $\wedge$-gates (resp., $\vee$-gates) of fan-in 2. \
After that, to enforce smoothness, for every disjunction gate (now with a fan-in 2) of subcircuits $C_1$ and $C_2$, find the set of variables in $C_1$, but not in $C_2$ (denoted $V_{1-2}$), along with those in $C_2$, but not in $C_1$ (denoted $V_{2-1}$). For every variable $v \in V_{2-1}$, redefine $C_1$ as $C_1 \wedge (v \vee -v)$. Similarly, for every variable $v \in V_{1-2}$,  redefine $C_2$ as $C_2 \wedge (v \vee -v)$. For example, for $(x_1 \wedge x_2 \wedge x_3) \vee (x_2 \wedge -x_3)$, becomes $((x_1 \wedge x_2) \wedge x_3) \vee ((x_2 \wedge -x_3) \wedge (x_1 \vee -x_1))$. \ This algorithm takes quadratic time in the size of the dDBC, which is  its number of edges \cite[sec. 3.1.2]{aaai21JMLR}, \cite{shi19}.

\begin{examplealt} \em (example \ref{ex:sdd} cont.) \ By interpreting decision points and elements as disjunctions and conjunctions, resp., the SDD in Figure \ref{fig:EjemploSDD}(a) can be easily converted into d-DNNF circuit. Only variables are affected by negations. \ Due to the children of node \raisebox{.5pt}{\textcircled{\raisebox{-.9pt} {{\footnotesize $3$}}}}, that do not have the same variables, the resulting dBBC is not smooth (but it has fan-in $2$). \ Algorithm \ref{alg:SDD2dDBC} transforms it into the dDBCSFi(2)
 in Figure \ref{fig:EjemplodDBC}.
\boxtheorem \end{examplealt}

\vspace{-13mm}
\section{\shap \ Computation: Experiments}\label{sec:comp}

\vspace{-2mm}The \href{https://www.kaggle.com/datasets/camnugent/california-housing-prices}{``California Housing Prices"} dataset  was used for our experiments (it can be downloaded  from Kaggle \cite{kaggle}).   It consists of 20,640 observations for 10 features with information on the block groups of houses in California, from the 1990 Census.  Table \ref{tab:dataset} lists and describes the features, and the way they are binarized, actually by considering
if the value is above the average or not.\footnote{Binarization could be achieved in other ways, depending on the feature, for better interaction with the feature independence assumption.} to better The categorical feature \#$1$ is one-hot encoded, giving rise to 5 binary features: \#$1_a$, ..., \#$1_e$.  Accordingly, we end up with 13 binary input features, plus the binary output feature, \#$10$, representing whether the median price at each block is high or low, i.e. above or below the average of the original \#$10$.
\ We used the  \href{https://www.tensorflow.org/}{``Tensorflow"} and \href{https://github.com/larq/larq}{``Larq"} Python libraries to train a BNN with one
 hidden layer, with as many neurons as predictors, i.e. 13, and one neuron for the output. For the hidden neurons, the activation functions are step function, as in (\ref{eq:thr}).\ignore{, with outputs $1$ or $-1$, whereas the step function for the output returns $1$ or $0$. All weights were rounded to binary values ($1$ or $-1$) and the biases were kept as real numbers. The loss function employed was the {\em binary cross-entropy}, defined by
$\nit{BCE}(\bar{y}, \hat{\bar{y}}) = -\frac{1}{|\bar{y}|} \sum_{i=1}^{|\bar{y}|} \left[ y_i \text{log}(\hat{y}_i) + (1-y_i) \text{log}(1-\hat{y}_i) \right]$,
where $\hat{\bar{y}}$ represents the labels predicted by the model and $\bar{y}$ are the true labels. \
The BNN ended up having a binary cross-entropy of 0.9041, and an accuracy of 0.6580, based on the test dataset.}

\vspace{-5mm}
\begin{table*}[h]
\caption{\label{tab:dataset}Features of the ``California Housing Prices" dataset.}\vspace{-2mm}
\begin{center}
    \setlength{\arrayrulewidth}{0.3mm}
    % \setlength{\tabcolsep}{8pt}
    % \renewcommand{\arraystretch}{1.0}
    % \hline
    \scriptsize
    \begin{tabular}{|p{0.7cm}|p{2.4cm}|p{3.0cm}|p{2.5cm}|p{2.8cm}|}
        \hline
        \bf{Id \#} &
        \bf{Feature Name} &
        \bf{Description} &
        \bf{Original Values} &
        \bf{Binarization}
        \\
        \hline
        % \specialrule{.25em}{0em}{0em}
        \#1 &
        \textit{ocean\_proximity} &  A label of the location of the house w.r.t sea/ocean
         &
        Labels \textit{1h\_ocean}, \textit{inland}, \textit{island}, \textit{near\_bay} and \textit{near\_ocean} &  Five features (one representing each label), for which $1$ means a match with the value of \textit{ocean\_proximity}, and $-1$ otherwise
         \\
        \hline
        \#2 &
        \textit{households} &  The total number of households (a group of people residing within a home unit) for a block
         &
        Integer numbers from $1$ to 6,082 &  $1$ (above average of the feature) or $-1$ (below average)
         \\
        \hline
        \#3 &
        \textit{housing\_median\_age} & The median age of a house within a block (lower numbers means newer buildings)
         &
        Integer numbers from $1$ to $52$ &  $1$ (above average of the feature) or $-1$ (below average)
         \\
        \hline
        \#4 &
        \textit{latitude} &  The angular measure of how far north a block is (the higher value, the farther north)
         &
        Real numbers from $32.54$ to $41.95$ &  $1$ (above average of the feature) or $-1$ (below average)
         \\
        \hline
        \#5 &
        \textit{longitude} &  The angular measure of how far west a block is (the higher value, the farther west)
         &
        Real numbers from $-124.35$ to $-114.31$ &  $1$ (above average of the feature) or $-1$ (below average)
         \\
        \hline
        \#6 &
        \textit{median\_income} &  The median income for households within a block (measured in tens of thousands of US dollars)
         &
        Real numbers from $0.50$ to $15.00$ &  $1$ (above average of the feature) or $-1$ (below average)
         \\
        \hline
        \#7 &
        \textit{population} &  The total number of people residing within a block
         &
        Integer numbers from $3$ to 35,682 &  $1$ (above average of the feature) or $-1$ (below average)
         \\
        \hline
        \#8 &
        \textit{total\_bedrooms} &  The total number of bedrooms within a block
         &
        Integer numbers from $1$ to 6,445 &  $1$ (above average of the feature) or $-1$ (below average)
         \\
        \hline
        \#9 &
        \textit{total\_rooms} &  The total number of rooms within a block
         &
        Integer numbers from $2$ to 39,320 &  $1$ (above average of the feature) or $-1$ (below average)
         \\
        \hline
    \end{tabular}
\end{center}
\end{table*}

\vspace{-8mm}
\begin{table*}[h]
\vspace*{-6.0mm}
\begin{center}
    \setlength{\arrayrulewidth}{0.3mm}
    %\setlength{\tabcolsep}{8pt}
    % \renewcommand{\arraystretch}{1.0}
    % \hline
    \scriptsize
    \begin{tabular}{|p{0.7cm}|p{2.4cm}|p{3.0cm}|p{2.5cm}|p{2.8cm}|}
        \hline
        \bf{Id \#} &
        \bf{Output} &
        \bf{Description} &
        \bf{Original Values} &
        \bf{Labels}
        \\
        \hline
        % \specialrule{.25em}{0em}{0em}
        \#10 &
        \textbf{\textit{median\_house\_value}} &  The median house value for households within a block (measured in US dollars)
         &
        Integer numbers from 14,999 to 500,001 &  $1$ (above average of the feature) or $0$ (below average)
         \\
        \hline
    \end{tabular}
\end{center} %\vspace{-8mm}
\end{table*}

\begin{figure*}[t]
\centering
\includegraphics[width=\textwidth]{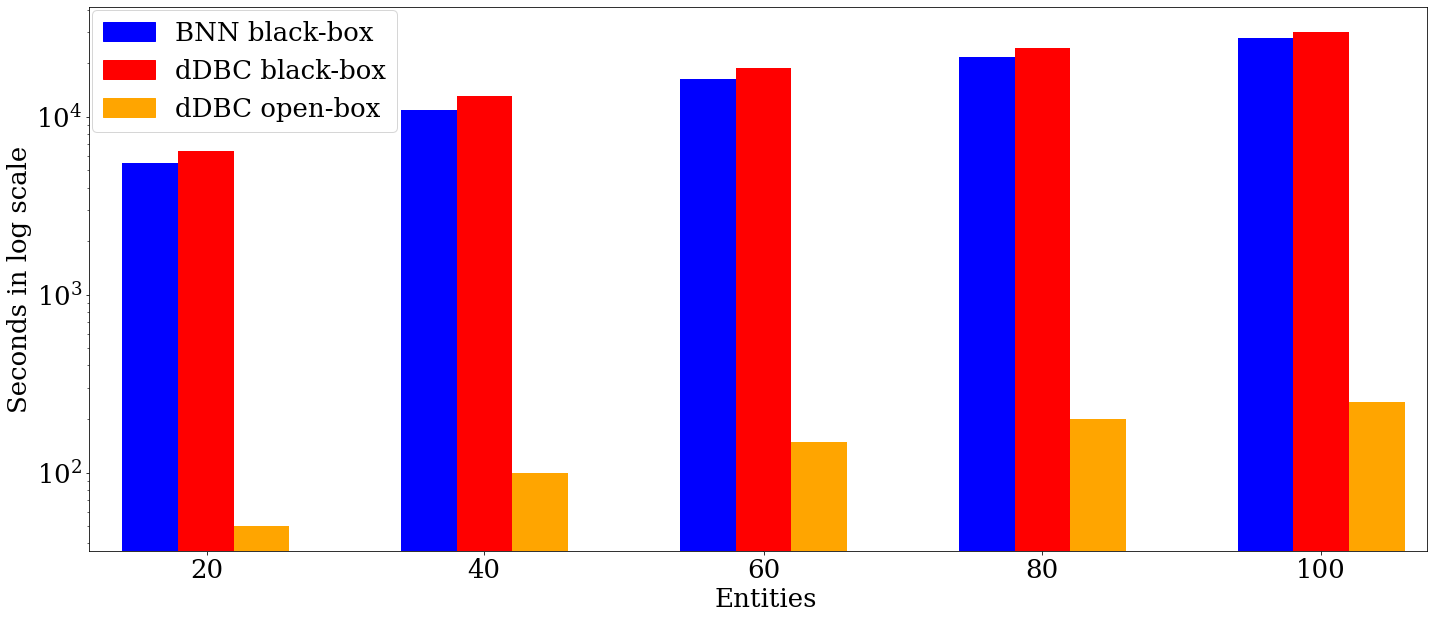}\vspace{-4mm}

\caption{Seconds taken to compute all \shap \ scores on 20, 40, 60, 80 and 100 input entities; using the BNN as a black-box (blue bar), the dDBC as a black-box (red bar), and the dDBC as an open-box (orange bar). Notice the logarithmic scale on the vertical axis.}\vspace{-5mm}
\label{fig:TiemposShap}\vspace{-1mm}
\end{figure*}

According to the transformation path (\ref{eq:path}), the constructed BNN was first represented as a CNF formula with 2,391 clauses. It has a tree-width of $12$, which makes sense having a middle layer of 13 gates, each with all features as inputs.   The CNF was transformed, via the SDD conversion, into  a dDBCSFi(2), $\mc{C}$, which ended up having 18,671 nodes (without counting the negations affecting only input gates). Both transformations were programmed in Python. For the intermediate simplification of the CNF, the {\em Riss} SAT solver was used \cite{riss}. The initial transformation into CNF took 1.3 hrs. This is the {\em practically} most expensive step, as was explained at the end of Section \ref{BNNtoCNF}. The conversion of the simplified CNF into the dDBCSFi(2) took 0.8276 secs.

With the resulting BC, we computed \shap, \ for each input entity, in three ways:\vspace{-1mm}
\begin{itemize}
    \item[(a)] Directly on the BNN  as a black-box model, using formula (\ref{eq:SHAPfunction}) and its input/output relation for multiple calls;
    \item[(b)] Similarly, using the circuit $\mc{C}$ as a black-box model; and
    \item[(c)] Using the efficient algorithm in \ignore{\cite{AAAI21}}\cite[page 18]{aaai21JMLR} treating circuit $\mc{C}$ as an open-box model. %For completeness, it is reproduced here as Algorithm \ref{alg:dDBCShap}.
\end{itemize}

\vspace{-1mm}These three computations of \shap \ scores were performed for sets of  20, 40, 60, 80, and 100 input entities,  for {\em all} 13 features, and {\em all} input entities in the set.  In all cases, using the uniform distribution over population  of size $2^{13}$. Since the dDBC faithfully represents the BNN, we  obtained exactly the same \shap \ scores under the modes of computation (a)-(c) above. The  {\em total} computation times were compared.   The results  are shown in Figure \ref{fig:TiemposShap}. \
Notice that these  times are represented  {\em in logarithmic scale}. \ For example, with the BNN, the computation time of all   \shap \  scores for 100 input entities was 7.7 hrs, whereas with the open-box dDBC it was 4.2 min. \ We observe a huge gain in performance with the use of the efficient algorithm on the open-box dDBC. Those times do not show the one-time computation for the  transformation  of the BNN into the dDBC. If the latter was added, each red and orange bar would have an increase of 1.3 hrs. For reference, even considering this extra one-time computation, with the open-box approach on the dDBC we can still  compute all of the \shap~scores for 100 input entities in less time than with the BNN with just 20 input entities.\footnote{The experiments were run on {\em Google Colab} (with an NVIDIA Tesla T4 enabled). Algorithm 1 was programmed in  Python. The complete code for {\em Google Colab} can be found at: \href{https://github.com/Jorvan758/dDBCSFi2}{https://github.com/Jorvan758/dDBCSFi2}.}

For the cases (a) and (b) above, i.e. computations with black-box models, the classification labels were first computed for all input entities in the population $\mc{E}$. Accordingly, when computing the \shap \ scores for a particular  input entity $\e$,  the labels for all the other entities related to it via a subset of features $S$ as specified by the game function were already  precomputed. This allows to compute formula (\ref{eq:SHAPfunction}) much more efficiently.\footnote{As done in \cite{deem}, but with only the entity sample.} The specialized algorithm for (c) does not require this precomputation. \ The difference in time between the BNN and the
black-box  dDBC, cases (a) and (b), is due the fact that BNNs allow some batch processing for the label precomputation\ignore{, which translates into  just 6 matrix operations  (2 weight multiplications, 2 bias additions, and 2 activations)}; with the  dDBC it has to be done one by one.

%\vspace*{-3mm}
\section{Conclusions}\label{sec:con}

\vspace{-1.5mm}We  have showed  in detail the practical use of logic-based knowledge compilation techniques in a real application scenario. Furthermore, we have applied them to the new and important problem of efficiently computing attribution scores for explainable ML. We have demonstrated the  huge computational gain, by comparing \shap \ computation with a BNN classifier   treated as an open-box vs.  treating it as  a black-box. The performance gain in \shap \ computation with the circuit exceeds by far both the compilation time and
the \shap \ computation time for the BNN as a black-box classifier.

We emphasize  that the effort invested in transforming the BNN into a dDBC is something we incur once.  The resulting circuit can be used to obtain \shap \ scores multiple times, and for multiple input entities. \ Furthermore, the  circuit can  be used for other purposes, such as {\em verification} of general properties of the classifier \cite{narodytska,darwicheEcai20}, and answering explanation queries about a classifier \cite{audemard}. \
Despite the intrinsic complexity involved, there is much room for improving the algorithmic and implementation aspects of the BNN compilation. The same applies to the implementation of the efficient \shap \ computation algorithm.

We computed \shap \ scores using the uniform distribution on the entity population. There are a few issues to discuss in this regard. First, it is computationally costly to use it with a large number of features. One could use instead the {\em empirical distribution} associated to the dataset, as in \cite{deem} for black-box \shap \ computation. This would require appropriately modifying the applied algorithm, which is left for future work. Secondly, and more generally, the uniform distribution does not capture possible dependencies among features. The algorithm is still efficient with the {\em product distribution}, which also suffers from imposing feature independence (see \cite{deem} for a discussion of its empirical version and related issues).   It would be interesting to explore to what extent other distributions could be used in combination with our efficient algorithm.

Independently from the algorithmic and implementation aspects of \shap \ computation, an important research problem is that of bringing {\em domain knowledge} or {\em domain semantics} into  attribution scores and their computations, to obtain more meaningful and interpretable results. This additional knowledge could come, for example, in  declarative terms, expressed as {\em logical constraints}. They could be used to appropriately modify the algorithm or the underlying distribution \cite{TPLP22}. It is likely that domain knowledge can be more easily be brought into a score computation when it is done on a BC classifier.

In this work we have considered only binary NNs. It remains to be  investigated to what extent our methods can be suitably modified for dealing with non-binary NNs.

\vspace{-2mm}
\paragraph{\bf Acknowledgments:} Special thanks to
Arthur Choi,
Andy Shih, Norbert Manthey,
Maximilian Schleich and  Adnan Darwiche, for their valuable help. \ Work was funded by ANID - Millennium Science
Initiative Program - Code ICN17002; CENIA, FB210017 (Financiamiento Basal para Centros Cient\'ificos y
Tecnol\'ogicos de Excelencia de ANID), Chile; SKEMA Business School, and NSERC-DG 2023-04650.  \ L. Bertossi is a Professor Emeritus at Carleton
University, Canada.

\newpage

\bibliographystyle{plain}

\end{document}